\documentclass{article}

\usepackage{microtype}
\usepackage{graphicx}
\usepackage{subfigure}
\usepackage{booktabs} 

\usepackage{hyperref}

\usepackage[accepted]{icml2025}

\usepackage{amsmath}
\usepackage{amssymb}
\usepackage{mathtools}
\usepackage{amsthm}

\usepackage[capitalize,noabbrev]{cleveref}

\theoremstyle{plain}

\theoremstyle{definition}

\theoremstyle{remark}

\usepackage[textsize=tiny]{todonotes}

\usepackage{makecell}
\usepackage{booktabs}
\usepackage{multirow} 
\renewcommand{\algorithmicrequire}{\textbf{Input:}}
\renewcommand{\algorithmicensure}{\textbf{Output:}}

\icmltitlerunning{How Effective Can Dropout Be in Multiple Instance Learning ?}

\begin{document}

\twocolumn[
\icmltitle{How Effective Can Dropout Be in Multiple Instance Learning ?}



\icmlsetsymbol{equal}{*}

\begin{icmlauthorlist}
\icmlauthor{Wenhui Zhu}{equal,asu}
\icmlauthor{Peijie Qiu}{equal,wasu}
\icmlauthor{Xiwen Chen}{equal,clemson}
\icmlauthor{Zhangsihao Yang}{asu}
\icmlauthor{Aristeidis Sotiras}{wasu}
\icmlauthor{Abolfazl Razi}{clemson}
\icmlauthor{Yalin Wang}{asu}
\end{icmlauthorlist}

\icmlaffiliation{clemson}{Clemson University, USA.}
\icmlaffiliation{wasu}{Washington University in St. Louis, USA.}
\icmlaffiliation{asu}{Arizona State University, USA.}

\icmlcorrespondingauthor{Wenhui Zhu}{wzhu59@asu.edu}
\icmlkeywords{Machine Learning, ICML}

\vskip 0.3in
]



\printAffiliationsAndNotice{\icmlEqualContribution} 

 \begin{abstract}
Multiple Instance Learning (MIL) is a popular weakly-supervised method for various applications, with a particular interest in histological whole slide image (WSI) classification. Due to the gigapixel resolution of WSI, applications of MIL in WSI typically necessitate a two-stage training scheme: first, extract features from the pre-trained backbone and then perform MIL aggregation. However, it is well-known that this suboptimal training scheme suffers from "noisy" feature embeddings from the backbone and inherent weak supervision, hindering MIL from learning rich and generalizable features. However, the most commonly used technique (i.e., dropout) for mitigating this issue has yet to be explored in MIL. In this paper, we empirically explore how effective the dropout can be in MIL. Interestingly, we observe that dropping the top-k most important instances within a bag leads to better performance and generalization even under noise attack. Based on this key observation, we propose a novel MIL-specific dropout method, termed MIL-Dropout, which systematically determines which instances to drop. Experiments on five MIL benchmark datasets and two WSI datasets demonstrate that MIL-Dropout boosts the performance of current MIL methods with a negligible computational cost. The code is available at \url{https://github.com/ChongQingNoSubway/MILDropout}.
\end{abstract}

\section{Introduction}
Multiple instance learning (MIL) has gained significant attention in machine learning applications. MIL assigns a single label (bag-level label) to a collection of instances (a bag), making it a weakly-supervised classification method due to the absence of instance-level labels. MIL has been applied to various fields~\cite{zhu2023self,chen2024timemil}, including digital pathology~\cite{ilse2018attention,li2021dual,shao2021transmil}, video analysis~\cite{babenko2010robust,quellec2017multiple}, and time series classification~\cite{early2023inherently,chen2024timemil}. In particular, MIL is a defacto standard for histological whole slide image (WSI) classification in digital pathology. This is because WSIs are gigapixel images, making them challenging to process using traditional machine learning or deep learning methods. 
\begin{figure}
    \centering
    \includegraphics[width=1.0\columnwidth]{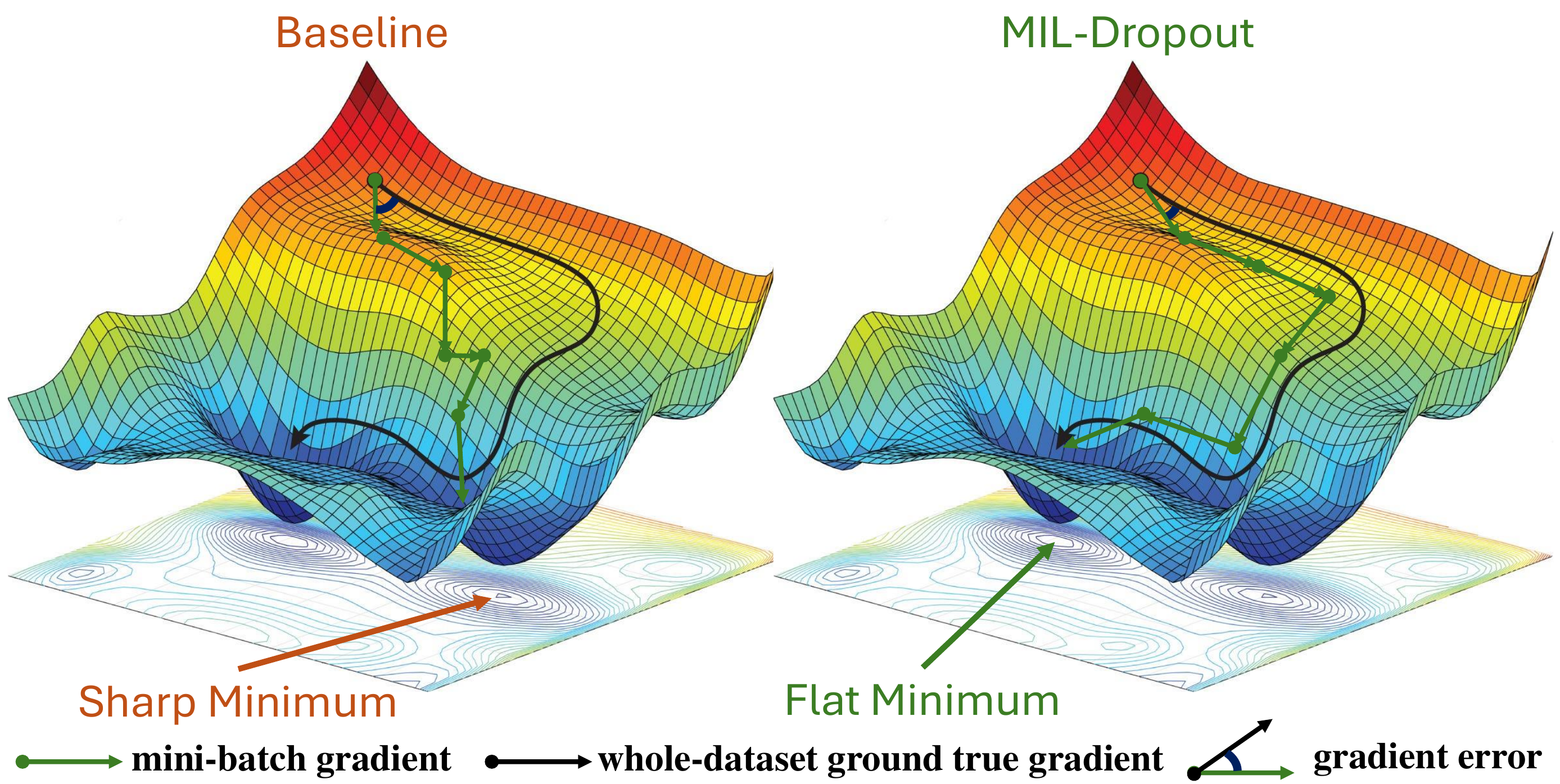}
    \caption{An illustrative example comparing the convergence trajectories of the baseline ABMIL without dropout (\textbf{Left}) and ABMIL with the proposed MIL-Dropout (\textbf{Right}). ABMIL without dropout is likely to follow an incorrect gradient direction initially and eventually converge to a sharp minimum. In contrast, ABMIL with the proposed MIL-Dropout typically achieves a lower gradient direction error and reaches a flatter minimum with a better generalization.}
    \label{fig:graphical}
\end{figure}
A WSI is treated as a bag comprising a collection of smaller tiles, each regarded as an instance. The MIL model takes a collection of WSI tiles as input and then assigns a single label to an entire WSI (e.g., malignant). Due to a large number of instances within a bag, deep learning based MIL methods~\cite{wang2018revisiting,ilse2018attention,shao2021transmil,zhang2022dtfd,zhu2024dgr,qiu2023sc} typically necessitate a two-stage learning scheme, where features are first extracted by a pre-trained backbone and then combined by an MIL aggregator. However, this suboptimal training scheme suffers from "noisy" feature embeddings, typically resulting in reduced performance. Follow-up methods also argue that this issue is attributed to overfitting and the inability of the MIL aggregator to learn a rich representation in such a suboptimal learning scheme~\cite{li2021dual}. Although it is not formally discussed,~\cite{li2021dual} randomly drops input instances (i.e., DropInstance in this paper) in their implementation, which partially evidences the effectiveness of this simple dropout-like method in regularizing MIL.
Motivated by this observation, we argue that a formal and thorough investigation of the role of the dropout method in MIL is necessary.


For this purpose, we first conduct experiments on various dropout strategies in MIL. Our empirical investigation indicates that dropping the top-k most important instances (i.e., top-k DropInstance) leads to an increase in classification accuracy. At first glance, this appears counterintuitive, whereas an in-depth analysis reveals its theoretically intriguing properties. First, Different from findings in other deep learning models~\cite{liu2023dropout,gradient1,gradient2}, dropping top-k most important instances can help MIL reduce the gradient direction error until converging (see example in Fig.~\ref{fig:graphical} and evidence in Fig.~\ref{fig:gde}). Second, it encourages the MIL model to converge to a flatter minimum, which enhances generalizability (e.g., robustness to noise attack). Despite its simplicity, dropping important instances not only results in performance gain but also shows theoretical guarantees. However, determining the importance of instances and selecting which ones to drop remains an open question.

Leveraging the advantageous properties of top-k DropInstance, we propose a novel MIL-Dropout method that systematically determines the importance of instances and identifies which ones to drop. This approach helps reduce gradient direction errors and achieves better generalization by converging to flatter local minima (see example in Fig~\ref{fig:graphical} (\textbf{Right}) and evidence in Fig.~\ref{fig:landscape}). Our method involves two key 
components: (i) a novel averaging-based attention mechanism to efficiently determine instance importance and (ii) a query-based instance selection mechanism to select the instance set to drop.
Extensive experiments on various MIL benchmarks demonstrate that our method can be seamlessly integrated into existing MIL frameworks to effectively improve their performance with negligible computational cost.

\section{Related Work}
\paragraph{MIL Methods.}
The introduction of MI-Net~\cite{wang2018revisiting} has significantly elevated the prominence of bag-level MIL methods, which utilize only bag-level labels for supervision. This approach addresses the ambiguity associated with propagating bag-level labels to individual instances inherent in instance-level MIL methods~\cite{feng2017deep, hou2016patch, xu2019camel}. Particularly in WSI classification, empirical studies have consistently demonstrated that bag-level MIL methods generally outperform their instance-level counterparts~\cite{shao2021transmil, wang2018revisiting}.
Recent advancements in bag-level MIL methods have primarily focused on improving instance-level MIL aggregation. This has been achieved through the incorporation of attention mechanisms~\cite{ilse2018attention}, transformers~\cite{shao2021transmil,xiang2023exploring}, pseudo bag~\cite{zhang2022dtfd}, and non-local attention~\cite{li2021dual} to effectively capture correlations between instances. In addition, some studies have begun focusing on negative mining within instances, using masked approaches to uncover more positive instances and incorporating a complex teacher-student model~\cite{milhard}. However, the rationale behind masking instances remains unclear, and these methods cannot be easily generalized to a wider range of MIL tasks.  In contrast, we focus on demonstrating the effectiveness of dropout in regularizing MIL to achieve better performance and generalization. Instead of enhancing MIL aggregation, we focus on integrating the proposed a general MIL-Dropout into various MIL aggregators to boost their performance.

\paragraph{Dropout methods.}
Dropout has been empirically shown to be an effective technique for mitigating overfitting in a variety of computer vision studies~\cite{dropout_naive, dropout}. Numerous follow-up studies have extended the concept of dropout, exploring a wide range of methodologies. Notable examples include spatial-dropout~\cite{dropout1d}, dropout based on contiguous regions~\cite{ghiasi2018dropblock}, and attention-based dropout~\cite{attdrop}. 
However, the application of dropout in MIL has not been explored.
In the context of MIL, where the MIL aggregator operates on instance embedding features devoid of spatial context, the dropouts based on spatial context are not applicable. Despite sharing similarities with the attention-based dropout, our MIL-Dropout is tailored to MIL applications, which drops instances instead of neurons in other applications (e.g., natural image classification). It is worth noting that this paper also serves as a theoretical supplement to PDL~\cite{zhu2023pdl}.



\section{Preliminary.}
The MIL is usually treated as a binary classification problem, the goal is to learn a direct mapping from a bag of $N$ instances $\boldsymbol{X}=\{\boldsymbol{x}_n \ | \ n = 1, \cdots, N\}$ to a binary label $Y \in \{0, 1\}$. In most real scenarios, the instance-level labels $\{y_n  \ | \ n = 1, \cdots, N\}$ are unknown, making it a weakly-supervised problem:
\begin{equation}
    Y = \begin{cases} 0, \ \text{iff} \  \sum_{n} y_n = 0, \\ 1, \ \text{otherwise}. \end{cases}
    \label{eqn:1}
\end{equation}

We considered embedding-based MIL as an example due to its popularity. The standard workflow of an embedding-based MIL involves (i) projecting instances into feature embeddings via an instance-level feature extractor and (ii) aggregating the instance-level features into a bag-level prediction through an MIL aggregator. Specifically, the instance-level feature extractor typically consists of a pretrained backbone feature extractor $f_{\texttt{backbone}}$ (e.g., ResNet) and a trainable shallow feature extractor $f_{\theta}$ parameterized by $\theta$ (e.g., multi-layer perception), due to the intractability to optimize the backbone~\cite{li2021dual}. The input instances $\boldsymbol{X}$ is projected into $D$-dimensional feature vectors $\boldsymbol{V}=\{\boldsymbol{v}_n \ | \ n = 1, \cdots, N\} \in \mathbb{R}^{N\times D}$ by applying the instance-level feature extractors: $\boldsymbol{V} = f_{\theta}(f_{\texttt{backbone}}(\boldsymbol{X}))$. The MIL aggregator is then applied to the instance-level feature embeddings to obtain the bag-level probabilistic prediction $\hat{\boldsymbol{Y}} \in [0, 1]$:
\begin{equation}
    \hat{\boldsymbol{Y}} = f_{\omega}(\rho_{\psi}(\{\boldsymbol{v}_n \ | \ n = 1, \cdots, N\})),
\end{equation}
where $f_{\omega}(\cdot)$ is a bag-level classifier parameterized by $\omega$, and $\rho_{\psi}$ is a permutation-invariant MIL pooling function parameterized by $\psi$. The learning of an MIL can be achieved by optimizing the binary cross-entropy loss over trainable parameters $\{\theta, \psi, \omega\}$.


Without loss of generality, we consider attention-based MIL pooling (ABMIL)~\cite{ilse2018attention} as an example. This is because most embedding-based MIL models~\cite{zhang2022dtfd,shao2021transmil} fall within the regime of ABMIL. Mathematically, an ABMIL pooling function is defined as
\begin{equation}
\begin{split}
     &\rho_{\psi}(\{ \boldsymbol{v}_n\ | \ n=1, \cdots, N \}) = \sum_{n=1}^{N} \alpha_n \boldsymbol{v}_n,  \\
     & \ \text{with} \ \ \alpha_n = \operatorname{softmax}(\boldsymbol{w}_1^T \operatorname{tanh}(\boldsymbol{w}_2 \boldsymbol{v}_n^T)), 
\end{split}
\label{eq:abmil-pooling}
\end{equation}
where $\psi= \{\textbf{w}_1 \in \mathbb{R}^{D \times 1}$, $\textbf{w}_2 \in \mathbb{R}^{D \times L} \}$ is the trainable parameter, and $\alpha_n$ implies the importance of the $n$-th instance.

\begin{figure}[!t]
    \centering
    \includegraphics[width=0.47\textwidth]{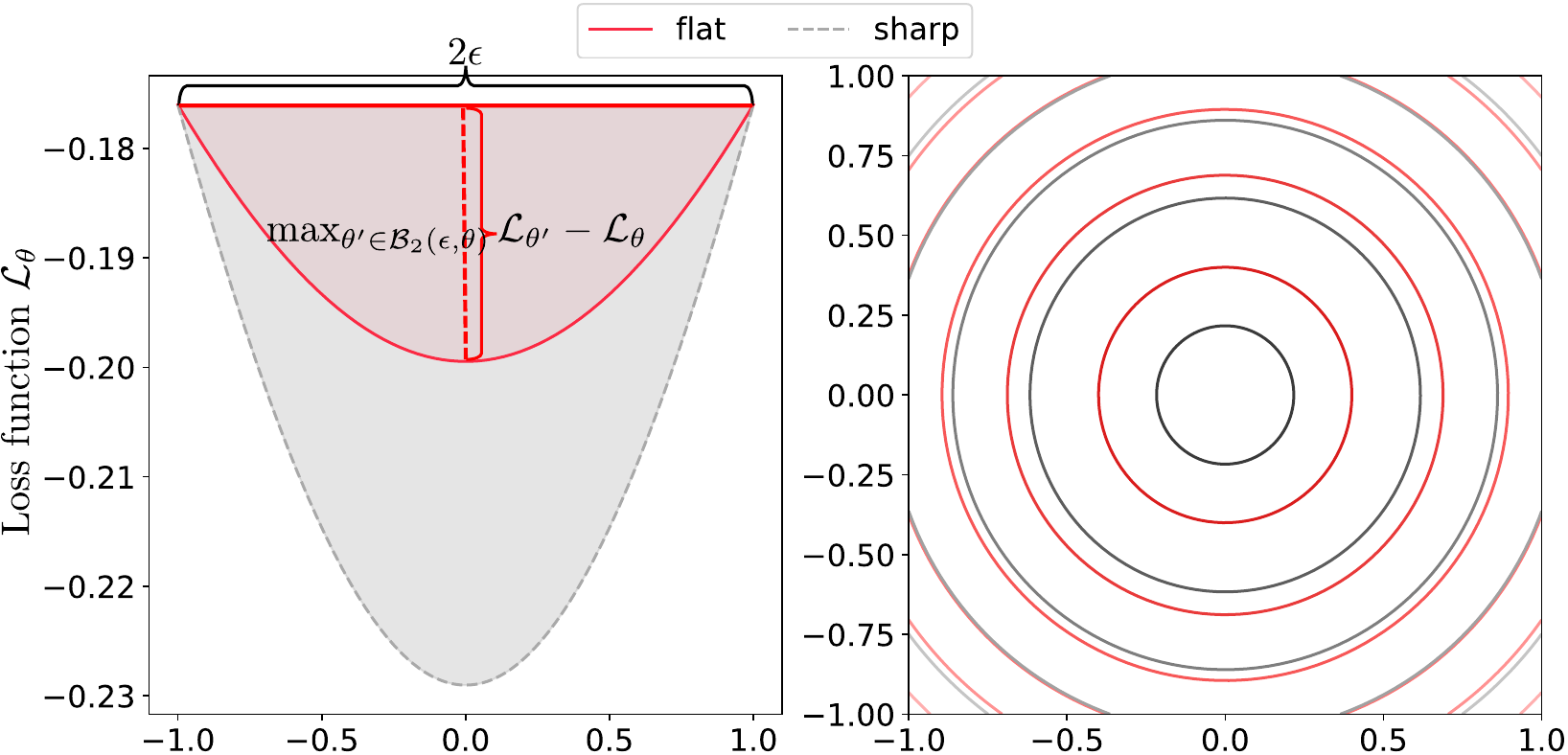}
    \caption{An conceptual illustration for a flat and sharp minimum in 1D curvature (\textbf{Left}) and 2D landscape (\textbf{Right}) of the loss function $\mathcal{L}_\theta$.}
    \label{fig:sharpness}
\end{figure}
\begin{figure*}[!t]
    \centering
    \includegraphics[width=0.99\textwidth]{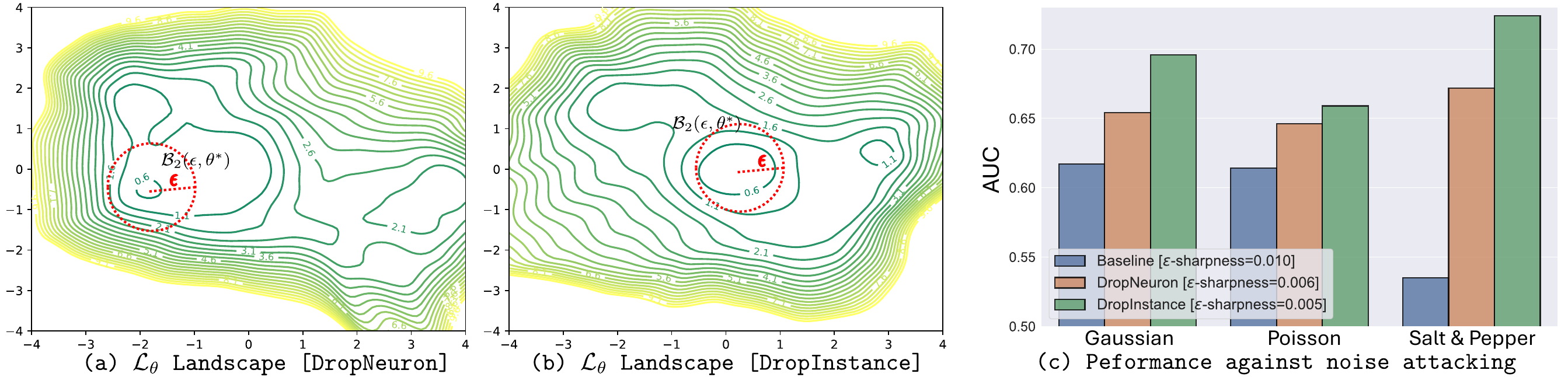}
    \caption{The landscape of the loss function $\mathcal{L}_{\theta}$ for two different dropout strategies (a) DropNeuron and (b) DropInstance as well as (c) the performance of MIL models against different noise attacks. We mark the Euclidean ball $\mathcal{B}_2(\epsilon, \theta^*)$ around the optimal parameter $\theta^*$ (see Eq. \ref{eq:flat1}) in subpanel figure (a) and (b) with a red circle. We observe that the landscape of $\mathcal{L}_\theta$ in the DropInstance scenario leads to flatter minima compared to DropNeuron, which also results in a better performance in AUC.}
    \label{fig:landscape}
\end{figure*}

\subsection{Analysis of Dropout in MIL}\label{sec:analysis}
In an MIL framework, Dropout is typically applied to the shallow feature extractor $f_{\theta}$. For simplicity, considering $f_{\theta}$ as an MLP with $L$ layers, the feature map at the $l$-th layer of $f_{\theta}$ is a 2-dimensional matrix $\boldsymbol{f}^{(l)} \in \mathbb{R}^{N \times D^{(l)}}$, where $D^{(l)}$ represents the embedding dimension. In this scenario, we consider randomly zeroing out either entries~\cite{dropout_naive} or entire instances~\cite{dropout1d} in $\boldsymbol{f}^{(l)}$.
For brevity, we denote the former as \emph{DropNeuron} and the latter as \emph{DropInstance}. For a given rate $p \in [0, 1]$, both DropNeuron and DropInstance can be defined as performing an element-wise masking operation over the feature map $\boldsymbol{f}^{(l)}$ at the $l$-th layer of $f_{\theta}$:
\begin{align}\label{eq:4}
    \hat{\boldsymbol{f}}^{(l)}= \boldsymbol{f}^{(l)} \odot \boldsymbol{M}^{(l)}, 
\end{align}
where $\odot$ denotes element-wise multiplication, and $\boldsymbol{M}^{(l)} \in \mathbb{R}^{N \times D^{(l)}}$ is the binary Dropout mask at the $l$-th layer. In the regime of DropNeuron, each entry $M^{(l)}_{n, d}$ in $\boldsymbol{M}^{(l)}$ are from a Bernoulli distribution: $M^{(l)}_{n, d} \sim Bernoulli(p)$. In contrast, each row in  $\boldsymbol{M}^{(l)}$ has the same entry and is sampled from a Bernoulli distribution in the case of DropInstance: $M^{(l)}_{n} \sim Bernoulli(p)$. 
However, how and when to impose Dropout remains an open question in the context of MIL, which has yet to be thoroughly explored by previous MIL works.
In this section, we provide insight into these two aspects by conducting empirical and theoretical analyses in the relatively complex WSI classification (e.g. Camelyon16) using a representative ABMIL framework. Please refer to Appendix A for the details of these investigation experiments. \emph{Our investigation reveals two main findings: (i) DropInstance leads to flatter local minima and better generalizability compared with DropNeuron, and (ii) DropInstance helps reduce gradient direction errors in learning trajectory, improving data fitting and performance.}

\subsection{DropNeuron vs DropInstance.}
A common way to assess the effectiveness of Dropout is to examine the robustness of an MIL after applying Dropout. For this purpose, we compute the sharpness of a MIL model~\cite{dinh2017sharp}, which pertains to how sensitive the model's performance is to changes in the model parameters. 
\begin{figure*}[!t]
\centering
\includegraphics[width=0.99\textwidth]{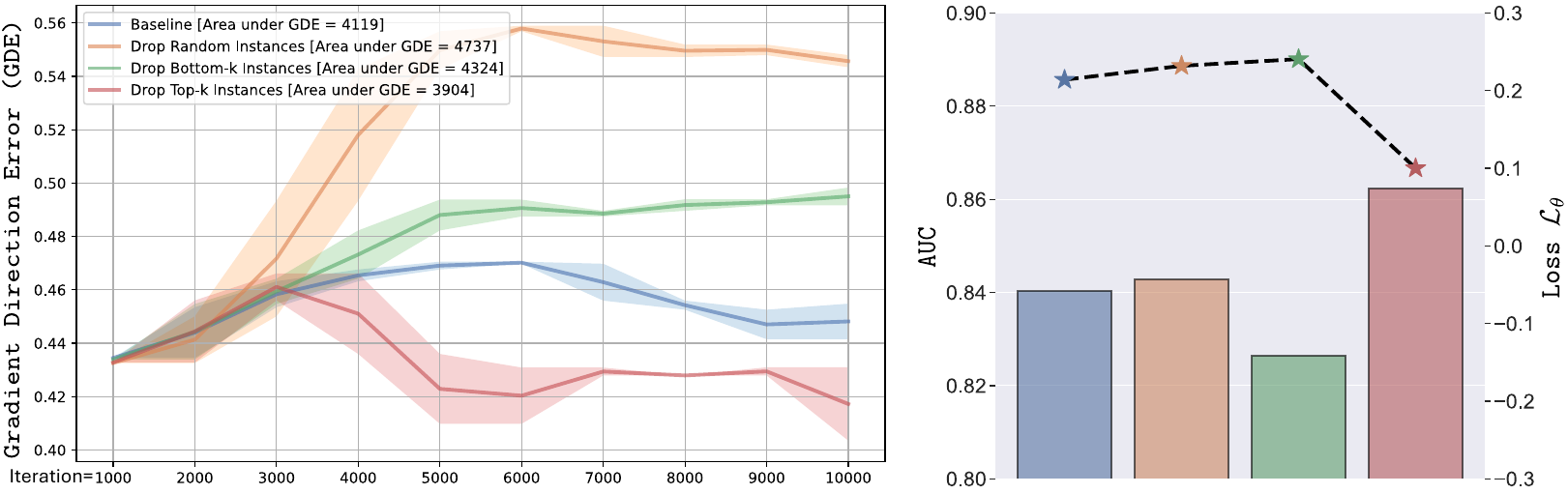} 
\caption{The comparison of change of GDE (\textbf{Left}) over the first 10,000 iterations as well as performance and loss (line plot) and AUC (bar plot) when using different instance dropout strategies (\textbf{Right}), where the area under GDE is the area enclosed by GDE and the x-axis. Dropping the top-k instances shows the smallest GDE, training loss , and highest AUC  among all four strategies. }
\label{fig:gde}
\end{figure*}
We consider the $\epsilon$-sharpness~\cite{dinh2017sharp} as the sharpness measure, which is defined using the local curvature of the loss function. 
Let $\mathcal{B}_2(\epsilon, \theta)$ be an Euclidean ball centered on a minimum $\theta$ with radius $\epsilon$. Then, for a non-negative valued loss function $\mathcal{L}_{\theta}$, the $\epsilon$-sharpness is proportional to the following quantity:
\begin{align}\label{eq:flat1}
\epsilon\text{-sharpness} \propto  \frac{\max _{\theta^{\prime} \in \mathcal{B}_2(\epsilon, \theta)}\mathcal{L}_{\theta^{\prime}}-\mathcal{L}_{\theta} }{1+\mathcal{L}_{\theta}}.
\end{align}
Geometrically, the $\epsilon$-sharpness is proportional to the height of the area enclosed by the curvature of $\mathcal{L}_{\theta}$ in Fig.~\ref{fig:sharpness}(\textbf{Left}). In the 2D landscape of $\mathcal{L}_\theta$ in Fig.~\ref{fig:sharpness}(\textbf{Right}), the loss function $\mathcal{L}_\theta$ changes more rapidly for sharper minima. As suggested by \cite{keskar2016large,foret2021sharpnessaware,zhang2024implicit}, for a specific $\epsilon$, a smaller volume enclosed by the curvature of $\mathcal{L}_{\theta}$ typically indicates flatter local minima (see Fig.~\ref{fig:sharpness}), which leads to better generalizability.
By applying the second-order Taylor approximation, $\epsilon$-sharpness in Eq.~\ref{eq:flat1} can be rewritten as  
\begin{align}
    \epsilon\text{-sharpness}  = \frac{\|\nabla^2_{\theta} \mathcal{L}_{\theta} \|_2 \epsilon^2}{2(1 +  \mathcal{L}_{\theta} )},
\end{align}
which enables us to directly compute the $\epsilon$-sharpness. 
We assess the robustness of the trained MIL model against three different types of noise (i.e., Gaussian, Poisson, and Salt \& Pepper noise) by attacking the input tiles/patches with them. The implementation of this experiment is detailed in Appendix A. We observe that adding Dropout lowers the $\epsilon$-sharpness value and results in flatter minima (Fig.~\ref{fig:landscape} (a) and (b)), which also improves the bag-level classification performance (Fig.~\ref{fig:landscape} (c)). In particular, adding DropInstance leads to the best performance. This is also supported by the sharpness analysis of the MIL model (see Fig.~\ref{fig:landscape} (b)), where DropInstance leads to flatter local minima that typically have better generalizability. We conjecture this may be because DropInstance is similar to data augmentation, which generates random instance combinations to improve the performance. However, this observation prompts a question: \textit{Are all instances worth being dropped out, or is the improvement due to specific combinations after DropInstances?}

\subsection{How to impose Dropout. }
Here, we further investigate how to apply DropInstance. Previous studies have revealed that the effectiveness of algorithms or modules (e.g.Dropout) can be reflected by the gradient direction error (GDE)  or gradient variance during model optimization~\cite{liu2023dropout,gradient1,gradient2}. The gradient direction error quantifies the dissimilarity between the mini-batch gradient $g_{step}$ and whole dataset gradient $\hat{g}$:

\begin{align}\nonumber
   \text{GDE} = \frac{1}{|G|} \sum_{g_{\text{step}} \in G} \frac{1}{2} \left( 1 - \frac{\langle g_{\text{step}}, \hat{g} \rangle}{\| g_{\text{step}} \|_2 \cdot \| \hat{g} \|_2} \right),
\end{align}
where $G$ is a set of mini-batch gradients. 
Leveraging GDE, we investigate the impact of three different DropInstance strategies, including dropping (i) top-k instances, (ii) bottom-k instances, and (iii) random instances. 
Here, top/bottom-k refers to dropping top/bottom $k$ most important instances based on the importance of instances (i.e., attention scores in any attention-based MIL framework) sorted in descending order. For this purpose, we plot the change of GDE as a function of training iterations. As shown in Fig.~\ref{fig:gde} (\textbf{Left}), the GDE of standard MIL does not decrease during model learning. Likewise, the GDE of the MIL models that apply random and bottom-k instance dropout even increases across training iterations. This suggests that the parameter updates are not moving in the correct direction, potentially compromising the fitting of the MIL model. Among the four aforementioned DropInstance strategies, only dropping the top-k instances leads to decreased GDE by model training. Further comparison of the aforementioned four DropInstance strategies in terms of classification performance suggests that a smaller GDE typically leads to better classification performance and lower training loss as shown in  Fig.~\ref{fig:gde} (\textbf{Right}). 

\subsection{Explanation.} In summary, empirical results show that dropping the top-k most important instances typically leads to better performance and gradient error direction. 
In the context of MIL, a positive bag is defined by the presence of at least one positive instance. However, in practice, the positive instance within each bag diversity leads many models to learn a single set of highly similar positive instances to make decisions. Dropping top-k instances deactivates these learned discriminative instances (zeroing them out in the feature map). With the bag label remaining unchanged, this forces the network to seek additional positive instance representations. From a gradient perspective, mini-batch gradients are generated by different sub-networks that are closer to the whole dataset gradient. This indicates that dropping these top-k instances allows the network to continue learning effectively and enhances performance, explaining why the top-k strategy results in lower gradient error.

\section{Proposed MIL-Dropout}
Contingent on the above discovery, we introduce a simple yet effective dropout method (termed MIL-Dropout), which can be easily plugged into existing MIL frameworks. MIL-Dropout mainly leverages the instance-based dropout operation to drop the top-k most important instances, as well as instances similar to them. The most important step revolves around selecting these instances in a systematic way. 
In a previous investigation, the attention map, which indicates the importance of each instance, is obtained from Eq.~(\ref{eq:abmil-pooling}). However, we argue that a key issue with this attention map is that it is produced by previous optimization iteration. Due to the instability of MIL training under weak supervision, top-k instance selection based on this attention map may introduce additional noise, especially at the very beginning of the training. In addition, the top-k most important instances at different layers of the feature extractor $f_\theta$ can potentially be different. To mitigate this issue, we propose a non-parametric attention mechanism to select the top-k instances for dropout. 
Specifically, we note that various methods for natural image processing employ average pooling to compress channel information and produce an attention map~\cite{cabm, park2018bam}. Inspired by this, we extend average pooling to our method to determine the importance of instances at the $l$-th layer of the feature extractor $f_{\theta}$, i.e., $\alpha_n^{(l)}$: 
 \begin{gather}
\begin{aligned} 
\alpha_n^{(l)} = & \ \operatorname{sigmoid}    \left( \{\operatorname{avgpool}( \boldsymbol{f}^{(l)}_n) \ | \ n =1, \cdots, N \}\right) \quad  \\  & \text{with} \ \operatorname{avgpool}( \boldsymbol{f}^{(l)}_n) = \frac{1}{D^{(l)}} \sum_{i=1}^{D^{(l)}} { \boldsymbol{f}^{(l)}_{n,i}}.
\end{aligned} \label{eqt:AAN} \raisetag{20pt} 
\end{gather}
Here, the average pooling is applied to each instance at embedding dimension $D^{(l)}$ after passing a $\operatorname{sigmoid}$ activation function to obtain the importance weight of each instance.  Based on the instance importance $\alpha_n^{(l)} $, we first select the top-k most important instances and then group the embedding features $\boldsymbol{f}^{(l)}$ at the $l$-th layer of $f_{\theta}$ into two groups: (i) top-k $\boldsymbol{f}^{(l)}_{T}$
and (ii) remaining feature $\boldsymbol{f}^{(l)}_{R}$.
 
We argue that simply dropping the top-k instances is insufficient. This is because instances within a bag may not be independent and identically distributed (i.i.d.) in most real-world applications~\cite{idd,ilse2018attention}. This is particularly true for WSI classification, where pathologists consider both the contextual information around a single area and the correlation information between different areas when making decisions. Based on these facts, we also drop instances that are highly similar to the selected top-k instances to further encourage the model to learn diverse representations. For this purpose,  we propose a query mechanism to select $G$ number of instances that is highly similar to the selected top-k instances from the remaining features $\boldsymbol{f}^{(l)}_{R}$ based on their similarity:
\begin{align}\label{eq:sim}
  S_{i,j} =   \frac{\boldsymbol{f}^{(l)}_{T,i}(\boldsymbol{f}^{(l)}_{R,j})^\top\quad }{\|\boldsymbol{f}^{(l)}_{T,i}\|_2\|\boldsymbol{f}^{(l)}_{R,j}\|_2}\quad \forall i\in [K], j \in [N-K], 
\end{align}
where $S_{i,j}$ denote the similarity score between $ \boldsymbol{f}^{(l)}_{T,i}$ and $ \boldsymbol{f}^{(l)}_{R,j}$ features, with a higher $S_{i,j}$ indicating higher similarity. 
The top-k instances are then combined with their $K\times G$ similar instances to form the final dropout instance set indexed by $\mathcal{A}$, containing $N_{\mathcal{A}}$ instances. Then the remaining instances are indexed by $\Bar{\mathcal{A}}$, containing $N_{\Bar{\mathcal{A}}}$ instances.   Following the convention in Eq.~(\ref{eq:4}), MIL-Dropout can be achieved by masking out the selected instance set $\mathcal{A}$:
 \begin{align}\label{eq:norm}
    &\quad \quad \Tilde{\boldsymbol{f}}^{(l)}_{\mathcal{A}} = \gamma ({\boldsymbol{f}^{(l)} } \odot \boldsymbol{M}) \\
     & \text{with} \quad \boldsymbol{M}_{\mathcal{A}} = 0 \ \text{and} \ \boldsymbol{M}_{\Bar{\mathcal{A}}} = 1.  \nonumber
\end{align}
 Here, we add a normalization term $\gamma =  N / (N-K(1 + G))$ to stabilize the training. Note that each top-k instance may have duplicate similar instances, necessitating deduplication. Therefore, the final number of dropout instances is typically less than $K+(K\times G)$. 
 Similar to other dropout methods~\cite{dropout,dropout_naive}, we do not apply MIL-Dropout during inference. The algorithmic taxonomy of the MIL-Dropout is shown in Algorithm~\ref{alg:1}.

\begin{table*}[t]
  \caption{Performance comparison on MIL benchmark datasets. Each experiment is performed five times with 10-fold cross-validation. We reported the mean of the classification accuracy ($\pm$ the standard deviation of the mean).}
  \centering
  \resizebox{0.70\textwidth}{!}{
  \begin{tabular}{l|ccccc}
    \toprule
    Methods     & MUSK1  & MUSK2 & FOX & TIGER & ELEPHANT \\
    \hline
    mi-Net & 0.889 $\pm$ 0.039  & 0.858 $\pm$ 0.049 & 0.613 $\pm$ 0.035 &  0.824 $\pm$ 0.034 &  0.858 $\pm$ 0.037   \\
    MI-Net &  0.887 $\pm$ 0.041 & 0.859 $\pm$ 0.046 & 0.622 $\pm$ 0.038 & 0.830 $\pm$ 0.032 & 0.862 $\pm$ 0.034       \\
    MI-Net with DS & 0.894 $\pm$ 0.042 & 0.874 $\pm$ 0.043 & 0.630 $\pm$ 0.037 & 0.845 $\pm$ 0.039 & 0.872 $\pm$ 0.032    \\
    MI-Net with RC & 0.898 $\pm$ 0.043 & 0.873 $\pm$ 0.044 & 0.619 $\pm$ 0.047 & 0.836 $\pm$ 0.037 & 0.857 $\pm$ 0.040    \\
    ABMIL &  0.892 $\pm$ 0.040 & 0.858 $\pm$ 0.048 & 0.615 $\pm$ 0.043 & 0.839 $\pm$ 0.022 & 0.868 $\pm$ 0.022   \\
    ABMIL-Gated & 0.900 $\pm$ 0.050 & 0.863 $\pm$ 0.042 & 0.603 $\pm$ 0.029 & 0.845 $\pm$ 0.018 & 0.857 $\pm$ 0.027   \\
    GNN-MIL & 0.917 $\pm$ 0.048 & 0.892 $\pm$ 0.011 & 0.679 $\pm$ 0.007 & 0.876 $\pm$ 0.015 & 0.903 $\pm$ 0.010   \\
    DP-MINN & 0.907 $\pm$ 0.036 & 0.926 $\pm$ 0.043 & 0.655 $\pm$ 0.052 & 0.897 $\pm$ 0.028 & 0.894 $\pm$ 0.030  \\
    NLMIL & 0.921 $\pm$ 0.017 & 0.910 $\pm$ 0.009 & 0.703 $\pm$ 0.035 & 0.857 $\pm$ 0.013 & 0.876 $\pm$ 0.011 \\
    ANLMIL & 0.912 $\pm$ 0.009 & 0.822 $\pm$ 0.084 & 0.643 $\pm$ 0.012 & 0.733 $\pm$ 0.068 & 0.883 $\pm$ 0.014 \\
    DSMIL & 0.932 $\pm$ 0.023  & 0.930 $\pm$ 0.020 & 0.729 $\pm$ 0.018 & 0.869 $\pm$ 0.008 & 0.925 $\pm$ 0.007 \\
    \hline
    \makecell[l]{ABMIL \\+ MIL-Dropout} & \underline{0.964 $\pm$ 0.033} & \underline{0.954 $\pm$ 0.019} & \textbf{0.789 $\pm$ 0.043} & \underline{0.917 $\pm$ 0.036}  & \textbf{0.934 $\pm$ 0.046}   \\
    \hline
    \makecell[l]{ABMIL-Gated \\+ MIL-Dropout}  & \textbf{0.967 $\pm$ 0.019} & \textbf{0.958 $\pm$ 0.021} & \underline{0.788 $\pm$ 0.016} & \textbf{0.919 $\pm$ 0.033}  & \underline{0.927 $\pm$ 0.033}  \\

    \bottomrule
  \end{tabular}
  }
  \label{tab:classic}
\end{table*}

\subsubsection{Complexity analysis.}
Our MIL-Dropout only adds additional two hyperparameters  (i.e., top-k number $K$ and similarity instance number $G$), without involving additional learnable parameters.
In each iteration, the main overhead is computing the mask, which is presented as,
\begin{align}\nonumber
    \underbrace{\mathcal{O}(N\log N)}_{\text{Top-k Sorting}}+ \underbrace{\mathcal{O}(K(N-K)D^{(l)})}_{\text{Compute Similarity}}+  \underbrace{\mathcal{O}(K(N-K)\log N)}_{\text{Subsequent Sorting}}.
\end{align}
The complexity is substantially reduced to the fact that $\mathcal{O}(N D^{(l)})$ due to $K$ is typically much smaller than $N$, e.g., $K=20, \text{whereas} \ N\approx 5000\sim 10000$. 

\section{Experimental designs}
 \subsection{MIL benchmarks.} The benchmark datasets include MUSK1, MUSK2, FOX, TIGER, and ELEPHANT, which are commonly used to evaluate and compare the performance of MIL algorithms. MUSK1 and MUSK2~\cite{dietterich1997solving} focus on molecule classification, with each bag containing instances representing atoms. Conversely, FOX, TIGER, and ELEPHANT~\cite{andrews2002support} involve image classification, where each bag represents images and contains instances that represent patches within those images. As consistent with the experimental protocols outlined in~\cite{li2021dual}, all experiments are conducted five times with a 10-fold cross-validation. We report the mean ($\pm$ std) for all MIL benchmark datasets. 

\subsection{WSI datasets.} The CAMELOYON16 dataset aims to identify metastatic breast cancer in lymph node tissue and consists of high-resolution digital WSIs. It is divided into a training set of 270 samples and a testing set of 129 samples. The TCGA-NSCLC dataset primarily identifies two subtypes of lung cancer: lung squamous cell carcinoma and lung adenocarcinoma. As outlined in~\cite{li2021dual}, 1037 WSIs were divided into a training set of 744 WSIs, a validation set of 83 WSIs, and a testing set of 210 WSIs. Following the threshold~\cite{li2021dual} and OTSU~\cite{zhang2022dtfd} preprocessing methods, each WSI was divided into non-overlapping $224\times224$ patches at a magnification of $\times20$. Two sets of patches were extracted using different frameworks: ResNet-50 from DTFD-MIL, yielding 1024-dimensional vectors per patch, and the SimCL contrastive learning framework from DSMIL, yielding 512-dimensional vectors per patch.

\begin{algorithm}[!t]
\caption{MIL-Dropout Mechanism}
\begin{algorithmic}[1]
\REQUIRE Input feature map $\boldsymbol{f}^{(l)}=\left\{ {\boldsymbol{v}}_{1}, \cdots,  {\boldsymbol{v}}_{N} \right\}$, $K$ and $G$
\ENSURE Processed Bag $\hat{\boldsymbol{f}^{(l)}}$ with MIL-Dropout
\STATE \textbf{Initial:} $M\leftarrow \boldsymbol{1}_{K\times D^{(l)}}$ (\textit{\# Initial mask})
\STATE Select top-k important instances: $( \boldsymbol{f}^{(l)}_T,   \boldsymbol{f}^{(l)}_R )\leftarrow \texttt{split}(\boldsymbol{f}^{(l)}), \quad \mathcal{A}\leftarrow  [K]$ \textit{(Eq. \ref{eqt:AAN})}
\STATE Compute the similarity matrix between rest instances $ \boldsymbol{f}^{(l)}_R$ and top-K instances $ \boldsymbol{f}^{(l)}_T$

(\textit{\# Obtain $G$ instances from $ {\boldsymbol{f}^{(l)}_R} $ that are most similar to every top-K instance})
\FOR{$i=1$ to $K$}
\STATE $A_i = \arg\max_{S \subseteq R, |S| = G} \sum_{j \in S}  S_{i,j}$ \textit{(Eq. \ref{eq:sim})}
\STATE $\mathcal{A}\leftarrow \mathcal{A} \cup A_i$
\ENDFOR
\STATE $M[\mathcal{A},:] = 0$
\STATE $ \hat{\boldsymbol{f}^{(l)}}\leftarrow\gamma ( {\boldsymbol{f}^{(l)}} \odot M)$ \textit{(Eq. \ref{eq:norm})}
\STATE \textbf{return} $  \hat{\boldsymbol{f}^{(l)}}$ (\textit{\# Masking and normalization})
\end{algorithmic}
\label{alg:1}
\end{algorithm}

\subsection{Baselines.} On five MIL benchmark datasets, we mainly plug MIL-Dropout into ABMIL as a proof of concept, while comparing with other baseline methods: mi-Net, MI-Net, MI-Net with DS, MI-Net wth RC~\cite{wang2018revisiting}, ABMIL, ABMIL-Gated~\cite{ilse2018attention}, GNN-MIL~\cite{tu2019multiple}, DP-MINN~\cite{yan2018deep}, NLMIL~\cite{wang2018non}, ANLMIL~\cite{zhu2019asymmetric}, and DSMIL~\cite{li2021dual}. On the WSI datasets, we plug MIL-Dropout into four state-of-the-art MIL aggregators and their variants to valid the effectiveness of the proposed method, i.e., ABMIL~\cite{ilse2018attention}, DSMIL~\cite{li2021dual}, TransMIL~\cite{shao2021transmil}, and DTFD-MIL~\cite{zhang2022dtfd}.

\subsection{Evaluation metrics and implementation details.}
For the MIL benchmark dataset, we performed five times with a 10-fold cross-validation for each method and reported the mean of the classification accuracy and the standard deviation. For two WSI datasets, we report the classification mean of accuracy, F1, and AUC were reported with standard deviation based on running five times of each experiment without fixing random seed. All baselines were implemented with the parameter configurations specified in their original papers. To incorporate the MIL dropout, we employed a unified specific architecture involving adding three fully connected layers as the shallow feature extractor $f_{\theta}$ before entering MIL aggregation. Furthermore, We reported all detailed experiment settings in Appendix B.

\begin{figure*}[!t]
\centering
\includegraphics[width=0.98\textwidth]{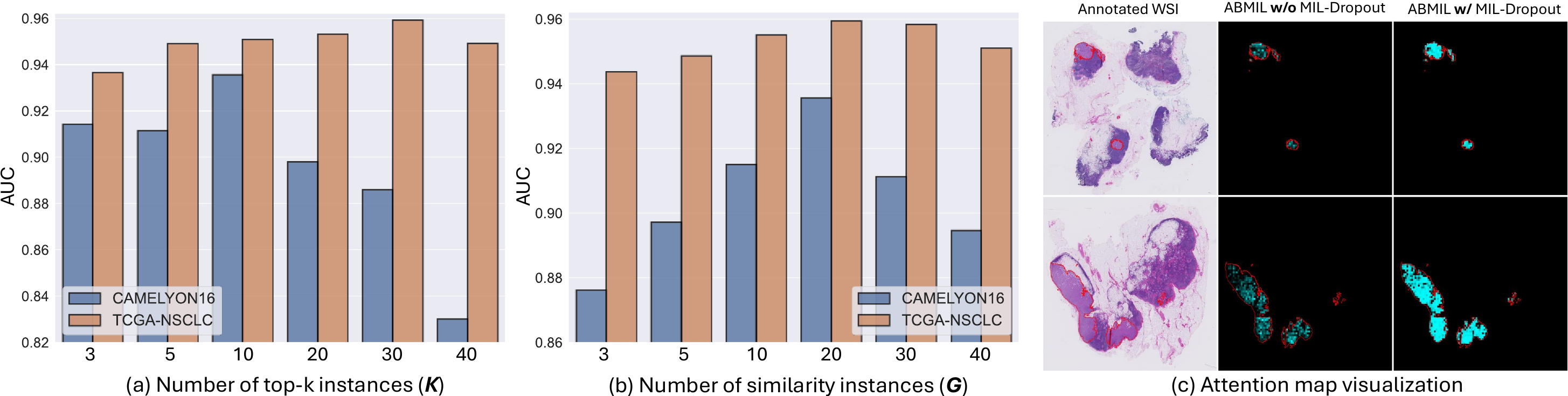} 
\caption{Ablation studies on the number of top-k instances $K$ (a) and similarity instance $S$ (b) using CAMELYON16 and TCGA-NSCLC datasets. (c) Attention map from ABMIL without and with MIL-Dropout, with tumor regions outlined in red. Brighter cyan in columns two and three indicates higher tumor probability (higher attention score) for corresponding locations.
}
\label{fig:abandlocalization}
\vspace{-0.2cm}
\end{figure*}

\begin{table*}[t]
\caption{Comparison of performance before and after plugging MIL-Dropout into four different types of MIL aggregators and their variants on CAMELOYON16 and TCGA-NSCLC datasets. 
$\Delta$ denotes the  performance gains after the integration of MIL-Dropout.
The classification accuracy (\%), F1 score (\%), and AUC (\%) are reported (± the standard deviation of the mean) by running each experiment five times. }
\centering
     \resizebox{1\linewidth}{!}{
\begin{tabular}{p{1.3cm}ccccccc|cccccc}
\toprule %
 & & \multicolumn{6}{c}{CAMELOYON16} & \multicolumn{6}{c}{TCGA-NSCLC} \\
 \cmidrule(r){3-8} \cmidrule(r){9-14}
 & & \multicolumn{3}{c}{ImageNet Pretrained} & \multicolumn{3}{c}{ SimCLR Pretrained} & \multicolumn{3}{c}{ ImageNet Pretrained} & \multicolumn{3}{c}{ SimCLR Pretrained} \\
 \cmidrule(r){3-5} \cmidrule(r){6-8} \cmidrule(r){9-11}  \cmidrule(r){12-14}
 & & Accuracy & F1 & AUC  & Accuracy & F1 & AUC & Accuracy  & F1 & AUC & Accuracy  & F1 & AUC \\
\toprule %

ABMIL & \makecell{\\+MIL Dropout \\ $\Delta$} & \makecell{86.3
 ± 1.1
  \\ 87.2 ± 1.0
 \\  \textbf{+0.9}} & 
\makecell{85.0±1.0  \\ 86.4±0.8
 \\  \textbf{+1.4}} & 
\makecell{86.0 ± 0.5 \\ 90.1± 0.8 \\ \textbf{+4.1}} & 

\makecell{85.6 ± 0.9 \\ 88.6 ± 1.1 \\ \textbf{+3.0}} & 
\makecell{84.2 ± 1.3 \\ 87.4 ± 1.0 \\  \textbf{+3.2}} & 
\makecell{ 86.6 ± 1.4 \\ 88.3 ± 1.2 \\ \textbf{+1.7}}

& \makecell{87.5 ± 0.8 \\ 91.1 ± 1.3 \\  \textbf{+3.6}} & 
\makecell{87.5 ± 0.8  \\ 91.1 ± 1.3 \\  \textbf{+3.6}} & 
\makecell{92.4 ± 0.5 \\ 95.6 ± 0.4 \\ \textbf{+3.2}} & 

\makecell{87.9 ± 0.8 \\ 91.4 ± 0.6 \\ \textbf{+3.5}} & 
\makecell{88.1 ± 0.6 \\ 91.5 ± 0.5 \\  \textbf{+3.4}} & 
\makecell{93.8  ± 0.8 \\ 95.9  ±  0.1 \\ \textbf{+2.1}}

\\ \cmidrule (l ){1 -14}

ABMIL-Gated & \makecell{\\+MIL Dropout \\ $\Delta$} & \makecell{86.9
 ± 1.1
  \\ 90.4 ± 1.3
 \\  \textbf{+3.5}} & 
\makecell{85.7±1.2  \\ 89.6±1.2
 \\  \textbf{+3.9}} & 
\makecell{86.2 ± 1.2 \\ 90.7 ± 0.9 \\ \textbf{+4.6}} & 

\makecell{84.3 ± 1.1 \\ 87.7 ± 1.3 \\ \textbf{+3.4}} & 
\makecell{83.4 ± 1.0 \\ 86.7 ± 1.3 \\  \textbf{+4.2}} & 
\makecell{ 85.9 ± 1.6 \\ 87.4 ± 0.9 \\ \textbf{+1.5}}

& \makecell{87.9 ± 0.9 \\ 90.0 ± 0.6 \\  \textbf{+2.1}} & 
\makecell{87.9± 0.9  \\ 90.0 ± 0.6 \\  \textbf{+2.1}} & 
\makecell{92.8 ± 0.9 \\ 95.3 ± 0.3 \\ \textbf{+2.5}} & 

\makecell{89.0 ± 1.2 \\ 90.8 ± 0.7 \\ \textbf{+1.8}} & 
\makecell{89.0 ± 1.2 \\ 90.8 ± 0.7 \\  \textbf{+1.8}} & 
\makecell{94.4 ± 0.7 \\ 95.8 ± 0.2 \\ \textbf{+1.4}}

\\ \cmidrule (l ){1 -14}

 DSMIL & \makecell{\\+MIL Dropout \\ $\Delta$} & 
 \makecell{85.5± 0.8\\ 87.9± 1.5\\ \textbf{+2.4}} & 
 \makecell{84.3± 1.1 \\ 86.8±1.6 \\  \textbf{+2.6}} & 
 \makecell{ 89.0± 1.8 \\ 90.6± 1.2\\ \textbf{+1.6}} & 
 
 \makecell{83.3 ± 1.0 \\ 85.6 ± 0.9 \\ \textbf{+2.3}} & 
 \makecell{82.0 ± 1.4 \\ 84.8 ± 0.5 \\  \textbf{+2.8}} & 
 \makecell{85.9 ± 1.6 \\ 87.6 ± 0.8 \\ \textbf{+1.7}}
 
 & \makecell{89.3 ± 0.7 \\ 89.9 ± 0.6 \\  \textbf{+0.6}} & 
 \makecell{89.4± 0.7  \\ 90.0± 0.5 \\  \textbf{+0.6}} & 
 \makecell{94.2 ± 0.3 \\ 95.3 ± 0.6 \\ \textbf{+1.1}} & 
 
 \makecell{84.1 ± 1.8 \\ 86.9 ± 0.4 \\ \textbf{+2.8}} & 
 \makecell{86.2 ± 1.5 \\ 88.3 ± 0.2 \\  \textbf{+2.1}} & 
 \makecell{92.0 ± 1.6 \\ 93.9 ± 0.3 \\ \textbf{+1.9}}
 
 \\ \cmidrule (l ){1 -14}

 TransMIL & \makecell{\\+ MIL Dropout \\ $\Delta$} &  \makecell{84.7 ± 2.1 \\  86.0 ± 1.5  \\ \textbf{+1.3}}  &  
 \makecell{83.3±2.9  \\ 84.9±1.5 \\  \textbf{+1.3}} & 
 \makecell{86.5  ± 2.4 \\ 89.4 ± 0.9 \\ \textbf{+2.9}}  &  
 
 \makecell{86.8 ± 1.0 \\ 89.7 ± 1.3 \\ \textbf{+2.9}} & 
 \makecell{85.9 ± 1.2 \\ 88.7 ± 1.4 \\  \textbf{+2.8}} & 
 \makecell{89.7 ± 0.6 \\ 90.3 ± 1.2 \\ \textbf{+0.6}}
 
 & \makecell{86.9 ± 0.6\\ 88.0 ± 0.5 \\  \textbf{+1.1}} & 
 \makecell{87.0 ± 0.6 \\ 88.5 ± 1.1 \\  \textbf{+1.6}} & 
 \makecell{93.3 ± 0.7 \\ 94.3 ± 0.4 \\ \textbf{+1.0}} & 
 
 \makecell{88.2 ± 2.1  \\ 91.6 ± 0.9 \\ \textbf{+2.8}} & 
 \makecell{88.3 ± 2.1  \\ 92.0 ± 0.7 \\  \textbf{+3.7}} & 
 \makecell{94.6 ± 1.1 \\ 96.2 ± 0.6 \\ \textbf{+1.6}}
 
 \\ \cmidrule (l ){1 -14}

DTFD-MIL(AFS) & \makecell{\\+MIL Dropout \\ $\Delta$} & \makecell{84.1 ± 0.6 \\ 85.7 ± 1.4 \\ \textbf{+1.6}} &  
\makecell{75.5 ± 0.6 \\ 79.1 ± 2.2 \\ \textbf{+3.6}} & 
\makecell{88.2 ± 0.3 \\ 89.9 ± 0.6 \\ \textbf{+1.6	}} & 

\makecell{87.4 ± 0.9 \\ 88.5 ± 0.7 \\  \textbf{+1.5}} & 
\makecell{81.8 ± 1.2 \\ 84.2 ± 0.6 \\ \textbf{+2.4}}&
\makecell{89.6 ± 0.9 \\ 92.5 ± 1.0 \\  \textbf{+2.9}} 

& \makecell{88.5 ± 0.5 \\ 90.3 ± 0.4 \\  \textbf{+1.8}} & 
\makecell{88.0 ± 0.3 \\ 90.0 ± 0.4\\  \textbf{+2.0}} & 
\makecell{94.4 ± 0.2 \\ 94.8 ± 0.1 \\ \textbf{+0.4}} & 

\makecell{87.6 ± 0.3 \\ 91.5 ± 0.4 \\ \textbf{+3.9}} & 
\makecell{87.8 ± 0.4 \\ 91.8 ± 0.4 \\  \textbf{+4.0}} & 
\makecell{ 93.1 ± 0.2 \\ 96.1 ± 0.2 \\ \textbf{+3.0}}

 \\ \cmidrule (l ){1 -14}

DTFD-MIL(MaxS) & \makecell{\\+MIL Dropout \\ $\Delta$} & \makecell{84.7 ± 1.8	 \\ 86.5 ± 0.9 \\ \textbf{+1.8}} &  
\makecell{78.3 ± 2.4 \\ 81.0 ± 1.3 \\ \textbf{+2.7}} & 
\makecell{87.8 ± 0.8 \\ 89.8 ± 0.9 \\ \textbf{+2.0}} & 

\makecell{87.7 ± 1.5 \\ 89.5 ± 0.4\\  \textbf{+1.8}} & 
\makecell{82.0 ± 2.3 \\ 84.4 ± 0.4 \\ \textbf{+2.4}}&
\makecell{88.4 ± 0.9 \\ 91.6 ± 0.5\\  \textbf{+3.2}} 

& \makecell{87.4 ± 1.0 \\ 88.8 ± 0.5 \\  \textbf{+1.4}} & 
\makecell{87.3± 1.0  \\ 88.4 ± 0.5 \\  \textbf{+1.1}} & 
\makecell{93.8 ± 0.1 \\ 95.0 ± 0.4 \\ \textbf{+1.2}} & 

\makecell{85.1 ± 1.2 \\ 87.5 ± 2.5 \\ \textbf{+2.4}} & 
\makecell{84.9 ± 2.2 \\ 88.2 ± 2.2 \\  \textbf{+3.3}} & 
\makecell{91.0 ± 1.0 \\ 93.2 ± 1.0 \\ \textbf{+2.2}}
\\

\bottomrule
\end{tabular}
}
\label{tab:wsi}
\end{table*}
\section{Results}
\subsection{MIL benchmark results.}
The integration of the proposed MIL-Dropout into ABMIL and ABMIL-Gated leads to superior performance compared to all prior state-of-the-art (SOTA) methods across five MIL benchmark datasets, as shown in Table~\ref{tab:classic}. Notably, ABMIL augmented with MIL-Dropout achieves SOTA accuracy of 78.9\% on the FOX dataset and 93.4\% on the ELEPHANT dataset, along with an average accuracy improvement of 9.72\% across the five datasets compared to the baseline ABMIL. Similarly, ABMIL-Gated with Dropout reaches SOTA accuracy of 96.7\% on MUSK1, 95.8\% on MUSK2, and 91.9\% on TIGER, demonstrating an average accuracy improvement of 9.82\% across the five datasets over the baseline ABMIL-Gate. Through extensive experiments, we observe that integrating our proposed Dropout into the two simplest bags embedding MIL aggregation (ABMIL and ABMIL-Gated) can also achieve SOTA performance.

\subsection{WSI results.}
As shown in Table~\ref{tab:wsi}, we perform a comparative study to evaluate the effectiveness of integrating the proposed MIL-Dropout into current MIL aggregators on CAMELYON16 and TCGA-NSCLC WSI datasets.   Experimental results show that MIL-Dropout boosts the performance of four different types of MIL aggregators across these two WSI datasets with features extracted using different means. Specifically, applying MIL-Dropout results in average improvements of 2.3\% in accuracy, 2.46\% in F1 score, and 2.48\% in AUC,  respectively, on the CAMELYON16 dataset. Similarly, on the TCGA-NSCLC dataset, MIL-Dropout leads to average increases of 2.05\% in accuracy, 2.44\% in F1 score, and 1.80\% in AUC. Remarkably, integrating MIL-Dropout into simple attention-based MIL aggregators (ABMIL and ABMIL-Gated) achieve a performance that is comparable to or even surpasses state-of-the-art methods. This shows that even ABMIL can significantly improve representation learning with MIL-Dropout, addressing underfitting and enhancing latent feature discovery.
Interestingly, we observe that MIL aggregators trained with features extracted using SimCLR outperforms that trained with features extracted by SimCLR on the CAMELYON16 dataset, but shows similar performance on the TCGA-NSCLC dataset. With the proposed MIL-Dropout, the performance of both ImageNet and self-supervised pre-trained backbone is largely indistinguishable under optimal conditions. On the TCGA-NSCLC dataset, both extractors result in similar performance, suggesting that more training samples improve the generalization of self-supervised feature extractors. However, our MIL-Dropout can enhance the generalization of different feature extractors without an additional self-supervised training stage.


\subsection{Ablation analysis.}
We perform ablation studies on two key hyperparameters in MIL-Dropout: the number of top-k instances ($K$) and the number of instances similar to the top-k instances ($G$). These analyses are carried out on the CAMELYON16 and TCGA-NSCLC datasets, using features extracted by ResNet-50 pretrained on ImageNet.
The optimal values for K varies from dataset to dataset, with $K=10$ on  CAMELYON16 and  $K=30$ on TCGA-NSCLC (Fig.~\ref{fig:abandlocalization}(a) and (b)). This suggests that TCGA-NSCLC may have large variability in instances. 
Additionally, a too-large K and G may disrupt MIL tasks by discarding too much information, whereas a too-small K and G may not be sufficient enough to drop easily identified positive instances and encourage MIL to discover richer features. As a result, we seek to find an equilibrium point that balances these effects.



\subsection{Lesion localization.}
As shown in Fig.~\ref{fig:abandlocalization} (c), ABMIL with MIL-Dropout provides better localization with a higher attention score to the positive instances compared to ABMIL without MIL-Dropout. In contrast, the attention map of vanilla ABMIL may either miss some true positive regions or show a lower importance. These results show evidence that MIL-Dropout can potentially help uncover more rich and generalizable lesion features,instead of relying on a single group of surely identified positive instances to make decisions. This aligns with the underlying rationale and theoretical guarantees behind the success of MIL-Dropout.

\section{Conclusion}
In this paper, we thoroughly analyze the role of dropout-like methods in MIL. Our observations empirically demonstrate that simply dropping certain instances can significantly enhance the performance and generalization of existing MIL methods. We further propose an MIL-Dropout to systematically determine the instances to drop by leveraging an attention-based mechanism to weigh instance importance and a query-based mechanism to select instance sets.
Experiments on several MIL datasets demonstrate the effectiveness of our MIL-Dropout. We hope our findings can guide the design of MIL models to learn more rich and generalizable features.




\section*{Acknowledgment}
This material is based upon the work supported by the National Science Foundation under Grant Number CNS-2204721. It is also supported by our collaborative project with MIT Lincoln Lab under Grant Numbers 2015887 and 7000612889. Additionally, this work was partially supported by National Institutes of Health, United States (R01EY032125 and R21AG065942) and the State of Arizona via the Arizona Alzheimer Consortium.

\section*{Impact Statement}
This paper proposes a novel dropout regularization method for multiple instance learning, derived from an empirical and theoretical study. The potential societal implications of MIL-Dropout can be categorized into three main areas:

\textbf{(i) Theoretical View.} Starting from a generalization perspective, we investigate two commonly used dropout methods in MIL—dropout and drop instance. We find that drop instance not only provides stronger generalization but also improves performance. Building on this, we analyze different instance-dropping strategies from an optimization standpoint and conclude that dropping the top-k instances achieves a lower gradient-direction error and reaches a flatter minimum with better generalization. Our findings and theoretical insights also fill an important gap for subsequent research on regularization in MIL.

\textbf{(ii) Applicability.} Our dropout method can be seamlessly integrated with existing MIL approaches and applied to all MIL algorithms, whether pathology or other weakly supervised classification tasks in natural images.


\textbf{(iii) Interpretability.} From Interpretability perspective, our method can enhance the localization capability in MIL-based weak supervision and help ensure model robustness. This, in turn, facilitates the identification of vulnerabilities and reduces the risk of adversarial attacks.

\textbf{Limitation.} At present, we have not extended this method to a broader range of tasks, such as video frame detection or 3D point cloud classification. Moreover, the current MIL dropout approach does not account for instance uncertainty, which could be a promising direction for guiding instance-drop strategies. We plan to explore these aspects in future work.

\nocite{langley00}

\bibliography{main}
\bibliographystyle{icml2025}

\newpage
\appendix
\onecolumn

\section{Investigation Experiment Design}
We implementation all experiments on a node of cluster with NVIDIA V100 (32GB). We use Pytorch Library \cite{paszke2019pytorch} with version of 1.13. The investigation experiments employ the same set of CAMELYON16 extractor features; for the pre-processing stage, we utilized Otsu's thresholding method to localize tissue regions within each slide image (WSI). Subsequently, non-overlapping patches, each measuring 256 × 256 pixels at 20X magnification, were extracted from these localized tissue regions. In total, 3.7 million patches were obtained from the CAMELYON-16 dataset~\cite{zhang2022dtfd}. Following the previous work~\cite{zhang2022dtfd}, The pre-trained Resnet-50 on ImageNet serves as the feature extractor for the embedding patch to a 1024-dimensional vector. The investigation experiments use the training/testing set provided by CAMELYON16 officials.

\begin{figure}
    \centering
    \includegraphics[width=0.99\columnwidth]{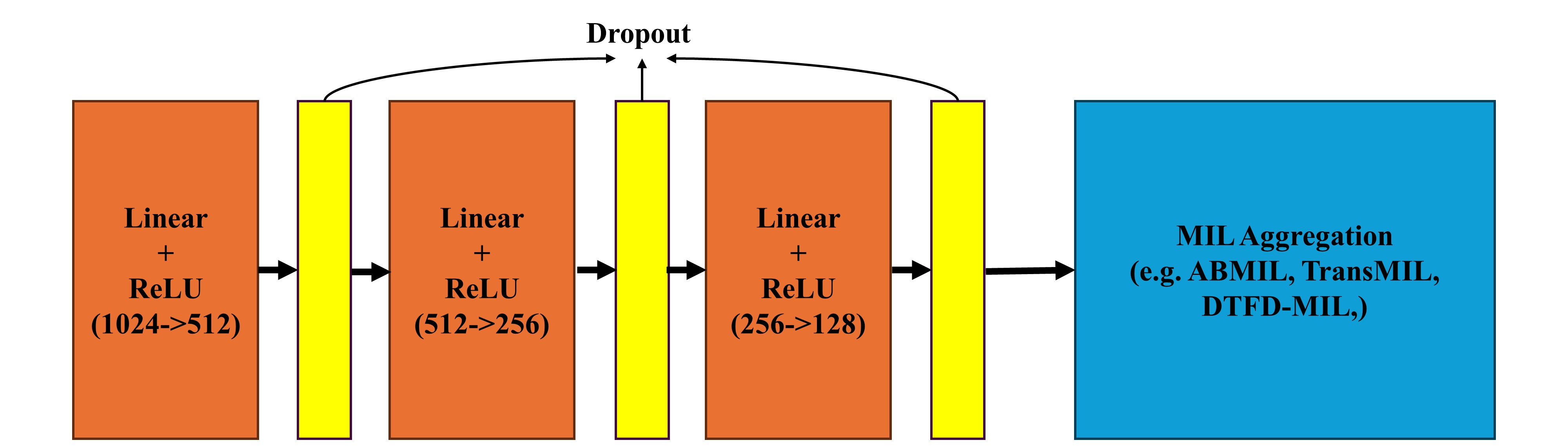}
    \caption{ABMIL aggregation with Dropout setting within investigation experiments .}
    \label{fig:appendix1}
\end{figure}

For these two experiments, we uniformly adopted the ABMIL~\cite{ilse2018attention} with cross-entropy loss, and the Lookahead optimizer was employed with a learning rate of 1e-4 and weight decay of 1e-4. To integrate Dropout into ABMIL, we employed three fully connected layers and ReLU before ABMIL aggregation(1024$\rightarrow$512$\rightarrow$256$\rightarrow$128). After applying the ReLU activation function to each layer, dropout regularization is introduced,as shown in Figure~\ref{fig:appendix1}.

\subsection{Detail of Generalization experiment}

\begin{figure}[htbp]
    \centering
    \includegraphics[width=0.99\columnwidth]{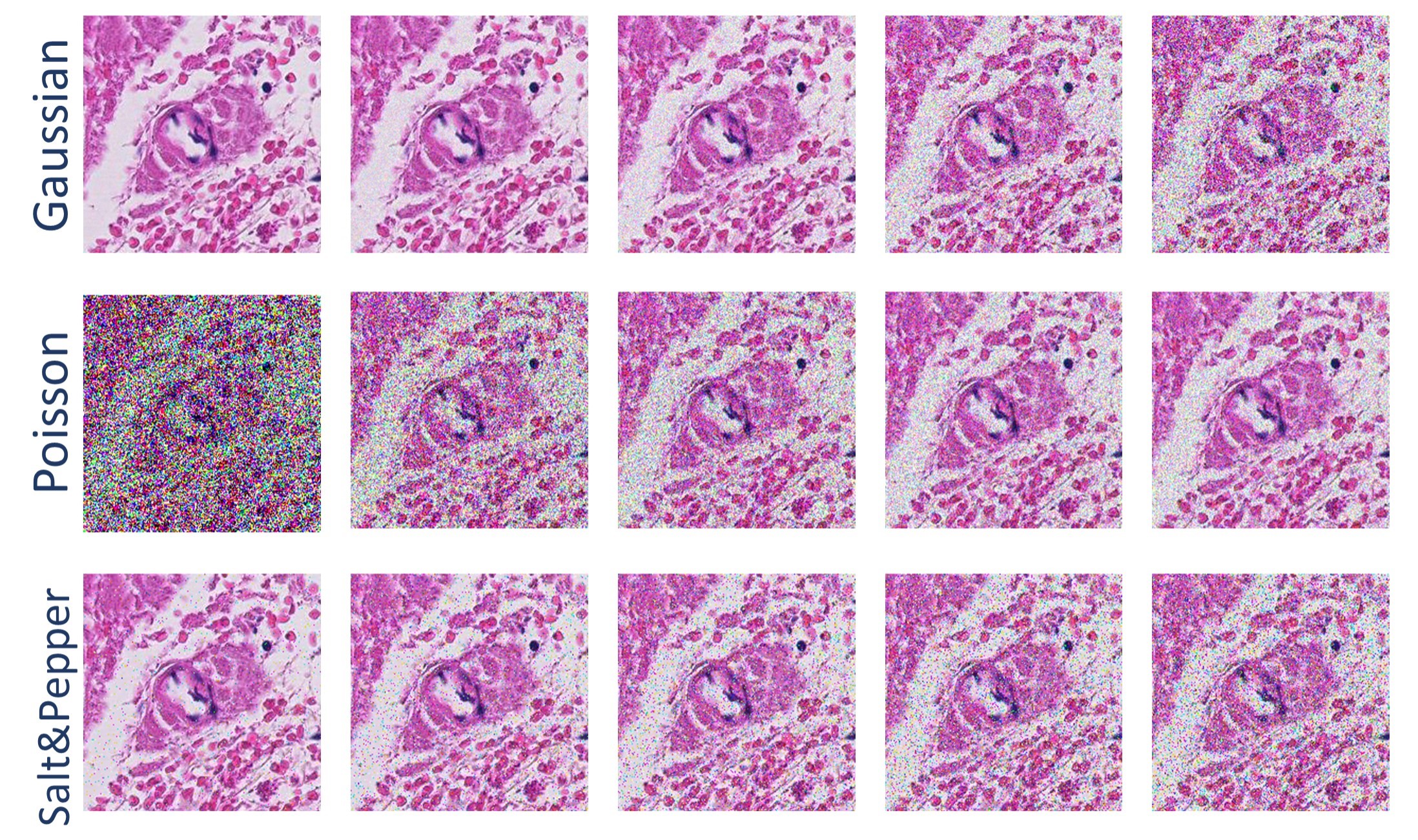}
    \caption{The exemplary patch with different noise types and strengths.}
    \label{fig:sup1}
    \vspace{-0.5cm}
\end{figure}

We conducted this experiment to investigate the CAMELYON16 dataset.
We train the model on the original training set and test its generalization on the test data with artifact noise to mimic the domain shift between training data and the unseen test set in real-world scenarios. We consider some common types of noise in the imagining process. Specifically, we add five levels for each type of noise: i) for Gaussian noise $\sigma\in\{0.01,0.05,0.1,0.2,0.3\}$, ii) for Poisson noise, we set its scale to $\{1.,5.,10.,20.,25.\}$, and iii) for Salt and Pepper noise, we set the probability of both salt and pepper to $\{0.01,0.03,0.05,0.07,0.09\}$. Therefore, we have 15 different noisy test sets now. Fig.\ref{fig:sup1} shows the exemplary patch with different noises. Apparently, a model with better generalization should be able to perform better on most of these datasets.

Now, let us elaborate on how to derive Eq. 5 to Eq. 6. The Taylor expansion for $L(\theta')$ presented in Eq. 5 can be expressed as,
\begin{align}
L(\theta') \approx L(\theta) +\nabla L(\theta)^\top (\theta' - \theta) + \nonumber 
+\frac{1}{2} (\theta' - \theta)^\top \nabla^2L(\theta) (\theta' - \theta) 
 +\mathcal{O}(\theta' - \theta).
\end{align}
Due to $\theta$ is a local minimal, $\nabla L(\theta)=0$, thereby,
\begin{align}
L(\theta')-L(\theta)=\frac{1}{2} (\theta' - \theta)^\top \nabla^2L(\theta) (\theta' - \theta) .
\end{align}
The maximum of a quadratic form occurs when $(\theta' - \theta)$ aligns with the eigenvector corresponding to the largest eigenvalue of $\nabla^2 L(\theta)$, therefore,
\begin{align}
    \frac{\max _{\theta^{\prime} \in B_2(\epsilon, \theta)}\left(L\left(\theta^{\prime}\right)-L(\theta)\right)}{1+L(\theta)} \nonumber
    = \frac{\epsilon^2 \| \nabla^2L(\theta)\|_2 }{2(1+L(\theta))},
\end{align}
where $\| \nabla^2L(\theta)\|_2$ denotes the spectral norm (largest eigenvalue).

\begin{table*}[t]
\centering
\caption{Parameters setting for baseline MIL methods. Here employed the AdamW, RAdam, CosineAnnealingLR to follow these MIL methods papers setting. }
\label{table2}
\resizebox{1\columnwidth}{!}{
\begin{tabular}{c|ccccc}
\toprule %
     & ABMIL & ABMIL-Gate & DSMIL & TransMIL & DTFD-MIL \\ \cmidrule (l ){1 -6}
    Optimizer & AdamW & AdamW &  AdamW & RAdam  & Adam \\ \cmidrule (l ){1 -6}
    Learning rate & $1e^{-4}$ & $1e^{-4}$&  $1e^{-4}$ & $1e^{-4}$ & $2e^{-4}$\\\cmidrule (l ){1 -6}
    Weight decay & $1e^{-4}$ & $1e^{-4}$ &  $5e^{-3}$ & $5e^{-3}$ &  $2e^{-3}$\\\cmidrule (l ){1 -6} 
    Optimizer scheduler & LookAhead & LookAhead &  CosineAnnealingLR & LookAhead  & MultiStepLR\\\cmidrule (l ){1 -6} 
    Loss function  & CrossEntropy & CrossEntropy &  CrossEntropy & CrossEntropy& CrossEntropy +  distille loss  \\
\bottomrule
\end{tabular}
}
\label{tb:parameters}
\end{table*}

\subsubsection{Grdient measure experiment}
We conducted this experiment based on the CAMELYON16 training and testing set. The primary objective of fitting the training data is to minimize the loss across the entire training set, rather than focusing on any individual mini-batch. Following in~\cite{liu2023dropout}, We compute the gradient for a given model over the entire training set, setting DropInstance to inference mode to capture the full model’s gradient. Subsequently, we evaluate the deviation of the actual mini-batch gradient $g_{step}$ from this whole-dataset "ground-truth" gradient $\hat{g}$. We define the average cosine distance from all 
 $g_{step}$ to $\hat{g}$ as the gradient direction error (GDE):
\begin{align}\nonumber
   \text{GDE} = \frac{1}{|G|} \sum_{g_{\text{step}} \in G} \frac{1}{2} \left( 1 - \frac{\langle g_{\text{step}}, \hat{g} \rangle}{\| g_{\text{step}} \|_2 \cdot \| \hat{g} \|_2} \right),
\end{align}
Here, we employ the same architecture as shown in Fig~\ref{fig:appendix1} for integrating Dropout. In this manuscript, we investigate four different DropInstance modes: top-k, bottom-k, random DropInstance, and without DropInstance. For the CAMELYON16 dataset, the value of $k$ is consistently set to 20. The terms "top" and "bottom" refer to the ABMIL aggregation attention map obtained from the previous iteration:
\begin{equation}
\begin{split}
     &\rho_{\psi}(\{ \boldsymbol{v}_n\ | \ n=1, \cdots, N \}) = \sum_{n=1}^{N} \alpha_n \boldsymbol{v}_n,  \\
     & \ \text{with} \ \ \alpha_n = \operatorname{softmax}(\boldsymbol{w}_1^T \operatorname{tanh}(\boldsymbol{w}_2 \boldsymbol{v}_n^T)), 
\end{split}
\label{eq:abmil-pooling}
\end{equation}
where $\alpha_n$ implies the importance of the $n$-th instance. In this context, the top and bottom refer to the ranking of attention scores in descending order, with the top corresponding to the highest attention scores and the bottom to the lowest. In addition, we conducted a performance comparison on the testing set. We observed that these models converge after approximately 10,000 iterations and achieve optimal performance prior to this point. In the manuscript, we present the highest AUC achieved before 10,000 iterations based on the four different DropInstance modes.

\section{Main Experiment Details}
We provide details of integrating MIL-Dropout into existing MIL methods, including ABMIL~\cite{ilse2018attention}, DTFD-MIL~\cite{zhang2022dtfd}, TransMIL~\cite{shao2021transmil}, DSMIL~\cite{li2021dual}.
\subsection{MIL Benchmark experiments setting}
\subsubsection{Embedding features} For the MUSK1 and MUSK2, A bag is constructed for each molecule, as instances. Each instance is represented with a 166-dimensional embedding feature. The FOX, TIGER, and ELEPHANT  datasets contain 200 bags of instance features. Each instance is associated with a 230-dimensional embedding feature.

\subsubsection{Detail of integrating MIL Dropout into existing MIL architectures} 
\begin{figure}
    \centering
    \includegraphics[width=0.99\columnwidth]{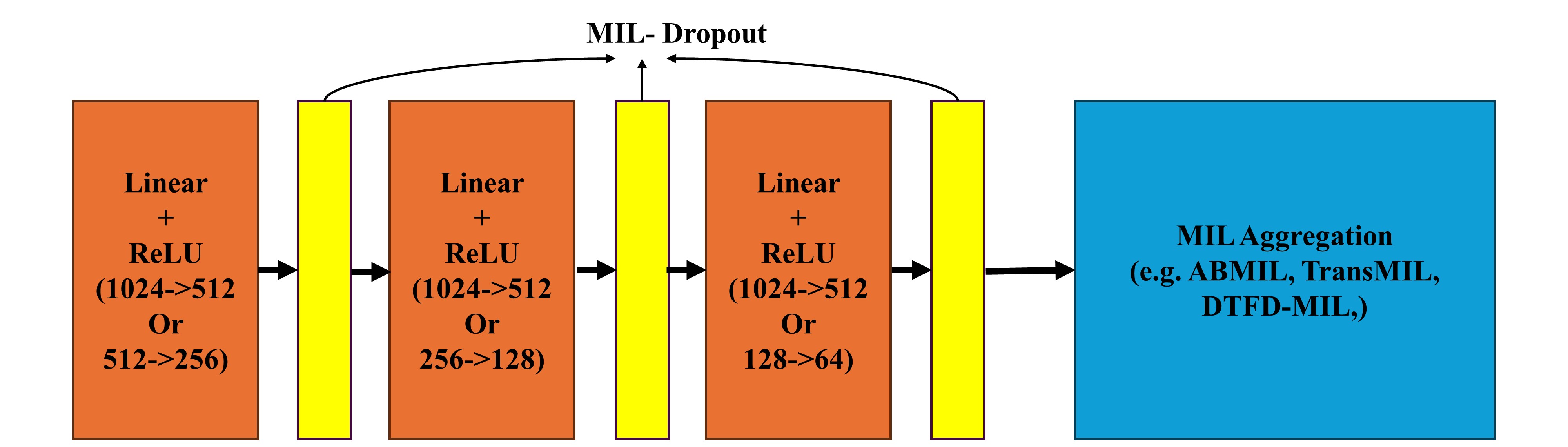}
    \caption{Different aggregation methods with MIL-Dropout setting within main experiments .}
    \label{fig:appendix2}
\end{figure}

We employed three fully connected layers in shallow feature extractor $f_{\theta}$, each comprising 256, 128, and 64 hidden units. Each layer was equipped with a ReLU activation function and subsequently appended by a MIL Dropout layer (refer to Fig~\ref{fig:appendix2}). Following the passage of features through the $f_{\theta}$, the ABMIL, ABMIL-Gate, and DSMIL aggregators processed the 64-dimensional embedding feature as their input. To compare original-based models, all aggregators adhere to the configuration outlined in paper~\cite{li2021dual}, with distinct embedding features of dimensions 230 and 166 directly employed as inputs. All experiments used the Adam optimizer with $2e^{-4}$ learning rates and $5e^{-3}$ weight decay and trained on cross-entropy loss for 40 epochs.

\subsection{WSIs experimental setting}
\subsubsection{Preprocessing WSI}  Following the threshold~\cite{li2021dual} and OTSU~\cite{zhang2022dtfd} preprocessing methods, each WSI was divided into non-overlapping $224\times224$ patches at a magnification of $\times20$. This results in a total of 3.7 million patches ($\sim 9000$ per bag) from the CAMELYON16 dataset and 8.3 million patches ($\sim 8000$ per bag) from the TCGA Lung Cancer dataset.


\subsubsection{Embedding Network pretrained}  Two sets of patches were extracted using different frameworks: \texttt{(i) ImageNet Pretrained:} ResNet-50 from DTFD-MIL, yielding 1024-dimensional vectors per patch, and \texttt{(i) SimCLR Pretrained:} the SimCL contrastive learning framework from DSMIL, yielding 512-dimensional vectors per patch. The self-supervised SimCLR manner employed the contrastive learning  framework~\cite{simplecontrastivelearning} to pretrain the projector based on the training set, wherein contrastive loss training was implemented between extracted patches and corresponding two random data data-augmentation counterparts~\cite{li2021dual}.
\subsubsection{Detail of integrating MIL Dropout into existing MIL architectures}
We followed the parameter settings outlined in the original literature for the baseline experiments on the two WSI datasets, as shown in Table~\ref{tb:parameters}. To integrate MIL Dropout into these MIL methods, we employed three fully connected layers in shallow feature extractor $f_{\theta}$, each comprising 512 / 256, 256 / 128, and 128 / 64 hidden units. Each layer was equipped with a ReLU activation function and subsequently appended by a MIL Dropout layer as shown in Fig~\ref{fig:appendix2}. The input dimensions of these networks were adjusted to 64 while keeping the other parameters unchanged. All the experiments were trained on 200 epochs. The MIL Dropout parameters were identical to the paper description (section Implementation details). The experiments integrated with MIL Dropout also followed the parameters in Table~\ref{tb:parameters}.

\section{Rebuttal Extra Experiment}
\subsection{UNI feature~\cite{chen2024uni}}
Consistent with requirements from other reviewers, we have replicated the relevant experiments here for clarity. Because UNI was pre-trained on TCGA and Camelyon data---raising the possibility of data leakage---we conducted additional experiments using the independent EBRAINS dataset. The results of these experiments are shown below.

\begin{table}[h!]
\centering
\begin{tabular}{lcccc}
\toprule
\textbf{Model} & \textbf{Accuracy} & \textbf{F1} & \textbf{$\Delta$ Accuracy} & \textbf{$\Delta$ F1} \\
\midrule
ABMIL & 65.4 & 68.7 & --- & --- \\
+MIL Dropout & 70.4 & 73.2 & +5.0 & +4.5 \\
TransMIL & 67.4 & 74.4 & --- & --- \\
+MIL Dropout & 71.3 & 79.4 & +3.9 & +5.0 \\
DSMIL & 67.4 & 74.4 & --- & --- \\
+MIL Dropout & 69.3 & 76.0 & +1.9 & +1.6 \\
DTFD & 53.4 & 63.6 & --- & --- \\
+MIL Dropout & 64.8 & 69.8 & +11.4 & +6.2 \\
\bottomrule
\end{tabular}
\end{table}

Although our MIL dropout can still offer improvements when the better patch features are implemented, the performance gains may be less substantial (about 1.5 to 0.5). Nevertheless, as demonstrated by the experiments, foundation model features often do not perform optimally on private datasets, making our approach particularly suitable for such scenarios.

\subsection{Additional Experiments}
Currently, our experiments primarily utilize shallow feature extractors. We also attempted to integrate MIL dropout into the classifier (add more linear layers for integrating MIL Dropout) following the MIL module (actually the same thing we felt), and the performance remained largely unchanged. Furthermore, integrating MIL dropout into a transformer architecture was unsuccessful---likely because MIL dropout interferes with layer normalization.

\begin{table}[h!]
\centering
\caption{Model (running 5 times on CAMELYON16 Imagenet)}
\begin{tabular}{lccc}
\toprule
\textbf{Model} & \textbf{Accuracy} & \textbf{F1} & \textbf{AUC} \\
\midrule
TransMIL & 84.7 & 83.3 & 86.5 \\
+MIL Dropout in shallow extractor & 86.0 & 84.9 & 89.4 \\
+MIL Dropout in classifier & 87.6 & 82.5 & 88.7 \\
+MIL Dropout in Transformer & 64.5 & 52.5 & 52.5 \\
\bottomrule
\end{tabular}
\end{table}

\textbf{MIL dropout fails to converge for integration into transformer blocks.}

\subsection{Extra Comparison}

Our MIL dropout can be seamlessly integrated with other MIL methods, as it is orthogonal to them. We conducted experiments using our patch features and observed performance improvements, further demonstrating the flexibility and effectiveness of our dropout.

\begin{table}[h!]
\centering
\caption{Model (running 3 times on CAMELYON16 Imagenet), [\cite{dpsf,qu2023boosting}]The full open-source code has not been released, and we believe our MIL dropout can be integrated into these methods.}
\begin{tabular}{lccccc}
\toprule
\textbf{Model} & \textbf{Accuracy} & \textbf{F1} & \textbf{AUC} & $\Delta$ \textbf{Accuracy} & $\Delta$ \textbf{AUC} \\
\midrule
PAM~\cite{pam} & 85.0 & 83.2 & 86.7 & --- & --- \\
+MIL Dropout & 86.2 & 84.3 & 87.7 & +1.2 & +1.0 \\
DPSF~\cite{dpsf} & --- & --- & --- & --- & --- \\
\cite{qu2023boosting} & --- & --- & --- & --- & --- \\
\bottomrule
\end{tabular}
\end{table}

\subsection{Questions/Suggestions}

\subsubsection{Experimental setup:}
Figure 5(c) reports ablation results on two datasets using ImageNet-pretrained features. All experiments in Table 2 use the optimal values of K and G identified in Figure 5(c).

\subsubsection{Performance degradation with high drop rate:}
\begin{itemize}
    \item Dropping a large proportion of instances inevitably degrades performance — not only for our proposed MIL-Dropout but also for standard/vanilla dropout.
    \item When most features or instances are zeroed out, the model struggles to distinguish positive from negative instances (ReLU outputs zero for both), while the bag-level label remains positive. This prevents effective learning.
    \item In our Camelyon16 experiments, aggressive dropout led to the failure of convergence for all methods; See the following results.
\end{itemize}

\begin{table}[h!]
\centering
\caption{Dropout Comparison on Camelyon16}
\begin{tabular}{lc}
\toprule
\textbf{Model} & \textbf{Accuracy} \\
\midrule
ABMIL & 86.3 \\
+Dropout(p=0.4) & 66.2 \\
+Dropout(p=0.8) & 53.4 \\
+Dropout1(D=0.4) & 63.1 \\
+Dropout1(D=0.8) & 51.3 \\
+Our MIL-Dropout (k = 100, G = 5) & 59.9 \\
+Our MIL-Dropout (k = 400, G = 5) & 55.7 \\
\bottomrule
\end{tabular}
\end{table}

\subsubsection{Optimal dropout threshold:}
\begin{itemize}
    \item By restricting dropout to at most 10\% of the average instance count (either via Top-K selection or a dropout probability $p = 0.1$), we observe consistent performance improvements across datasets.
\end{itemize}

\subsection{Extra WSI task/datasets Evaluation:}

\subsubsection*{Task \& Datasets:}
We evaluated our method on survival prediction using two TCGA datasets (LUAD and BRCA), following the protocols of \cite{jaume2024modeling, song2024multimodal}.

\subsubsection{Evaluation Metric:}
Concordance index (C-index) with \%. The K = 10, G = 10 for MIL-Dropout.

\subsubsection{Results:}
\begin{table}[h!]
\centering
\caption{Survival Prediction Results on TCGA-LUAD and TCGA-BRCA}
\begin{tabular}{lcccccc}
\toprule
\textbf{Model} & \textbf{TCGA-LUAD} & \textbf{+MIL-Dropout} & $\Delta$ & \textbf{TCGA-BRCA} & \textbf{+MIL-Dropout} & $\Delta$ \\
\midrule
ABMIL & 65.7 & 67.7 & +2.0 & 72.8 & 75.2 & +2.4 \\
DSMIL & 61.4 & 63.4 & +2.0 & 68.8 & 72.1 & +3.3 \\
TransMIL & 64.3 & 67.2 & +2.9 & 72.1 & 74.7 & +2.6 \\
DTFD (AFS) & 62.0 & 65.3 & +3.3 & 71.6 & 74.6 & +2.5 \\
DTFD (MaxS) & 65.9 & 68.8 & +2.4 & 72.8 & 75.7 & +2.9 \\
\bottomrule
\end{tabular}
\end{table}

\noindent
\textbf{These results demonstrate that our MIL Dropout consistently improves all baseline methods on WSI-based survival prediction.}

\end{document}